%% file: main.tex
\documentclass[journal,10pt,letterpaper]{IEEEtran}
\newif\ifnamedversion
\namedversiontrue
\usepackage[T1]{fontenc}
\usepackage[utf8]{inputenc}
\usepackage{amsmath,amssymb}
\usepackage{graphicx}
\usepackage{booktabs,array,tabularx}
\usepackage{cite}
\usepackage{tikz,pgfplots}
\usetikzlibrary{arrows.meta,positioning,calc,fit}
\pgfplotsset{compat=1.18}
\usepackage[protrusion=false]{microtype}
\usepackage[nocheckfootnote]{flushend}
\usepackage[hidelinks]{hyperref}
\usepackage{url}
\newcommand{\R}{\mathbb{R}}
\newcommand{\zh}{\widetilde{\hat z}}
\newcommand{\zt}{\tilde z}
\newcommand{\at}{\tilde a}

\newcommand{\rstate}{r_{\mathrm{state}}}
\newcommand{\rcmd}{r_{\mathrm{cmd}}}
\newcommand{\rspread}{r_{\mathrm{spread}}}
\newcommand{\dRMS}{d_{\mathrm{RMS}}}
\newcommand{\dmax}{d_{\max}}
\newcommand{\Qlin}{Q^{\mathrm{lin}}}
\newcommand{\calib}{\mathrm{cal}}
\newcommand{\tr}{\mathrm{tr}}
\newcommand{\sr}{\mathrm{sr}}
\newcommand{\norm}[1]{\lVert #1\rVert_2}
\newcolumntype{L}[1]{>{\raggedright\arraybackslash}p{#1}}
\newcolumntype{Y}{>{\raggedright\arraybackslash}X}
\hypersetup{pdftitle={Can Predicted Dynamics Exist in the Physical World? Diagnostics at the Prediction-Control Interface},pdfauthor={},pdfsubject={Runtime diagnostics for learned robot proposals},pdfkeywords={Anomaly detection, black-box models, learned dynamics, model diagnostics, runtime monitoring}}

\ifnamedversion
  \hypersetup{pdfauthor={Barak Or}}
\fi

\makeatletter
\let\STATEsixteenOriginalPlain\ps@plain
\def\ps@plain{%
  \STATEsixteenOriginalPlain
  \def\@oddhead{\hfil\normalfont\fontsize{9}{11}\selectfont\bfseries UNDER REVIEW\hfil}%
  \let\@evenhead\@oddhead
}
\makeatother
\hypersetup{pdfsubject={UNDER REVIEW - Runtime diagnostics for learned robot proposals}}
\title{\fontsize{23}{27}\selectfont Can Predicted Dynamics Exist in the Physical World?\\Diagnostics at the Prediction-Control Interface}
\ifnamedversion
  \author{Barak Or\thanks{Barak Or is with STATE16 (e-mail: \texttt{barak@state16.ai}).}}
\else
  \author{\mbox{}}
\fi
\begin{document}
\maketitle
\thispagestyle{plain}
\pagestyle{plain}
\begin{abstract}
Can learned state--action proposals exist in the physical world? Before executing a proposed action sequence, a robot can inspect empirical variation and disagreement with a learned transition model. However, aggregating these diagnostic signals obscures whether a proposal is inconsistent with the predictor or merely departs from recorded behavior. We formalize this prediction--control interface, separate the scalar trigger from its channel-wise diagnostic log, and establish that an all-pairs displacement term is redundant within a maximum that already contains the corresponding one-step term. We evaluate the monitors on 700 nominal and 5,250 synthetically perturbed 32-transition PushT windows, observing only planar pusher positions and goals. The transition-RMSE baseline achieves a ROC AUC of 0.982, compared with 0.957 for a heterogeneous maximum and 0.972 for a spread-scaled residual.
\end{abstract}
\section{Introduction}
Many learned robot policies emit short action sequences rather than one-step commands, and learned predictors can provide state forecasts over the same horizon~\cite{rt1,rt2,openx,octo,openvla,pi0}. These proposals can be inspected before execution, but the available monitoring signals have different meanings. A known platform limit, an unusually large sample-to-sample change, and disagreement with a learned transition model each support a different conclusion. Treating them as interchangeable obscures both the trigger rule and the information retained for later analysis.

Improved accuracy in learning-aided navigation~\cite{or2022hybrid,or2023trajectory,freydin2022speed} leaves a separate question: when can erroneous black-box outputs propagate into system failure? This motivates examining learned outputs before their use by downstream estimation or control.

Established safety filters, shielding methods, reachability analyses, and runtime-assurance architectures already mediate between planning and execution under explicit models and assumptions~\cite{ames,wabersich,hsu,simplex,shielding,hobbs}. CommonRoad checks trajectories against vehicle, road, and collision models~\cite{commonroad}; FORCE-OPT evaluates motion plans against calibrated reachable sets derived from trajectory predictions~\cite{forceopt}; and semigroup consistency tests whether a learned simulator agrees with its own composed evolution~\cite{shikhman}. Our setting uses a deliberately limited input interface: the monitor sees only a decoded rollout and empirical, model-relative scores.

We call this point the \emph{prediction--control interface} and study how its trigger score and channel-wise diagnostic log should be evaluated separately. Fig.~\ref{fig:interface} summarizes the setup. A trigger indicates that a selected score exceeds its reference threshold; it is not, by itself, a decision to authorize or block an action.

This paper makes three contributions. First, we distinguish platform constraints, empirical variation, transition-model disagreement, and predictor-interface consistency, and derive an explicit redundancy result for the all-pairs displacement term. Second, we define six controlled PushT perturbation operators and an episode-split comparison protocol. Third, we show in one fixed PushT run that the heterogeneous maximum has a lower ROC AUC than the transition-RMSE and spread-scaled-residual baselines.

\begin{figure*}[t]
\centering
\input{figures/interface.tex}
\caption{Monitoring at the prediction--control interface. A decoded proposal is passed to the diagnostic monitor before downstream use. The monitor returns two distinct outputs: a thresholded scalar score and a channel-wise log. Direct/composed forecasts are an optional interface diagnostic and are not evaluated by the fixed-horizon PushT predictor.}
\label{fig:interface}
\end{figure*}

Section~II reviews related work, Section~III defines the scores and trigger rule, Section~IV describes the PushT study, and Section~V reports the results. Sections~VI and VII discuss limitations and conclude.
\section{Literature Review}
\label{sec:related}
\textbf{Proposal-generating models.}
Action-chunking policies, diffusion policies, and VLA models generate temporally extended commands~\cite{diffusion,act,openvla,pi0}; learned transition models predict their consequences for planning or policy improvement~\cite{pets,mbpo,dreamer,dreamerv3}. What can be monitored depends on what the model exposes. A decoded rollout containing both states and actions permits cross-stream checks. An action-only chunk requires a separate predictor, and a latent plan must first be decoded. Unlike these proposal-generating methods, the present work studies how to interpret and compare the diagnostic signals available \emph{after} a proposal has been produced; it does not introduce another policy or predictor architecture.

\textbf{Learning-aided estimation.}
Prior work integrates learned noise covariances~\cite{or2022hybrid,or2023trajectory}, speed pseudomeasurements~\cite{freydin2022speed}, Kalman gains (KalmanNet~\cite{revach2022kalmannet}), or displacements and uncertainties (TLIO~\cite{liu2020tlio}) into state estimators. Our focus is on decoded state-action proposals: comparing trigger scores while separating transition disagreement from empirical variation.

\textbf{Existing interfaces between prediction and control.}
Safe control and runtime assurance use plant models, invariant sets, verified backup controllers, or explicit constraints~\cite{safeRL,ames,wabersich,simplex,hobbs}. CommonRoad checks collision avoidance, road compliance, and kinematic feasibility relative to a specified vehicle model and tolerances~\cite{commonroad}. FORCE-OPT constructs multimodal forward-reachable sets from trajectory predictions, calibrates their coverage under stated exchangeability assumptions, and tests planned trajectories for intersection with those sets~\cite{forceopt}. Our addition is narrower: we separate a model-relative trigger from its diagnostic log when only a decoded state--command sequence is available. No reachable-set, invariance, or physical-feasibility guarantee is claimed for this restricted interface.

\textbf{Predictor diagnostics.}
Ensemble disagreement is often used as a model-relative uncertainty proxy~\cite{ensembles}. For an autonomous learned simulator, Shikhman~\cite{shikhman} studies a structural diagnostic: direct evolution over $h+k$ steps should agree with evolution over $h$ followed by $k$ when the same predictor exposes all three calls. Building on these distinct interpretations, we keep residual, spread, and empirical-difference channels separate, identify an analytically redundant displacement channel, and test whether their heterogeneous maximum improves ranking over a simple transition residual.

\textbf{Direct versus composed prediction.}
Let $z\in\R^{d_z}$ be the initial observed state, with $d_z$ state coordinates, and let $a_i\in\R^{d_a}$ be command $i$, with $d_a$ command coordinates. The notation $a_{r:s}=(a_r,\ldots,a_s)$ includes both endpoints. The positive integers $h$, $k$, and $\ell$ denote horizons measured in samples. A learned predictor $\hat\phi_\ell$ maps an initial state and $\ell$ commands to the predicted state after $\ell$ steps. Define
\begin{subequations}\label{eq:composition-action}
\begin{align}
y_h &= \hat\phi_h(z,a_{0:h-1}),\label{eq:composition-intermediate}\\
y_{\mathrm{dir}} &= \hat\phi_{h+k}(z,a_{0:h+k-1}),\label{eq:composition-direct}\\
y_{\mathrm{comp}} &= \hat\phi_k(y_h,a_{h:h+k-1}),\label{eq:composition-composed}\\
\Delta^{F,a}_{h,k}(z,a_{0:h+k-1})
 &= \norm{y_{\mathrm{dir}}-y_{\mathrm{comp}}}.\label{eq:composition-residual}
\end{align}
\end{subequations}
Here $y_h$, $y_{\mathrm{dir}}$, and $y_{\mathrm{comp}}$ are, respectively, the intermediate, direct, and composed predicted states, all in $\R^{d_z}$; $\norm{\cdot}$ is the Euclidean norm. For an autonomous, state-only Markovian predictor, the corresponding quantity is
\begin{equation}
\Delta^F_{h,k}(z)=\norm{\hat\phi_{h+k}(z)-\hat\phi_k(\hat\phi_h(z))}.
\label{eq:composition-state}
\end{equation}
The superscript $F$ labels forecast-composition consistency and $a$ labels the action-conditioned case; neither is an exponent. A queried split fails this check when $\Delta^{F,a}_{h,k}>\varepsilon^{F,a}_{h,k}$, where $\varepsilon^{F,a}_{h,k}\geq0$ is a specified tolerance in the same coordinate units as the residual. The analogous state-only tolerance is $\varepsilon^F_{h,k}$. All compared calls must use the same state representation. These are predictor self-consistency diagnostics~\cite{shikhman}, not tests of a physical law. The direct PushT predictor considered in this work accepts exactly $K=32$ commands and is not queried at shorter horizons, so neither composition residual is measured in the study.
\section{Trigger and Diagnostic Log}
\label{sec:method}
Let $t$ be the current sample index and $K\in\mathbb{N}$ the proposal horizon. Local indices $i=0,\ldots,K$ are relative to $t$. A decoded proposal contains $K+1$ state entries $\hat z_i\in\R^{d_z}$ and $K$ commands $a_i\in\R^{d_a}$:
\begin{equation}
q_t=(\hat z_{0:K},a_{0:K-1}),\qquad \hat z_0\approx z_t.
\label{eq:proposal}
\end{equation}
For the higher-order differences below, assume $K\geq3$, as in the $K=32$ study. The monitor computes several diagnostic channels from the same proposal. Section~III-C explains how one scalar is used to trigger while all channels remain available for inspection.

\subsection{Empirical Variation Scores}
Let $\bar z_{\tr}$ and $s^z_{\tr}$ be the component-wise mean and population standard deviation of training-state entries. Define $\bar a_{\tr}$ and $s^a_{\tr}$ analogously for commands, and set $\epsilon=10^{-6}$. Standardize each coordinate:
\begin{equation}
\zt_j=\frac{z_j-\bar z_{\tr,j}}{s^z_{\tr,j}+\epsilon},\qquad
\at_j=\frac{a_j-\bar a_{\tr,j}}{s^a_{\tr,j}+\epsilon}.
\label{eq:standardize}
\end{equation}
The state expression uses $j=1,\ldots,d_z$ and the command expression uses $j=1,\ldots,d_a$. Applying the same transformation to every proposal entry gives $\zh_i$ and $\at_i$. For a vector sequence $x_{0:L}$, its forward difference of order $p$ is
\begin{equation}
\Delta^p x_i=\sum_{\ell=0}^{p}(-1)^{p-\ell}\binom{p}{\ell}x_{i+\ell},\quad 0\leq i\leq L-p.
\label{eq:forward}
\end{equation}
Here $L$ is the last sequence index and $\binom{p}{\ell}$ is the binomial coefficient.

The notation $\Qlin_\alpha$ denotes the empirical $\alpha$-quantile computed by linear interpolation. Let $\mathcal E_{\tr}$ index training episodes and $T_e$ be the number of commands in episode $e$. The collection $V^z_p$ contains $\norm{\Delta^p\zt_i^{(e)}}$ for every $e\in\mathcal E_{\tr}$ and valid index $0\leq i\leq T_e-p$. For $p\in\{1,2,3\}$, define
\begin{subequations}\label{eq:state}
\begin{align}
c^z_p&=1.25\,\Qlin_{0.995}(V^z_p)+\epsilon,\label{eq:state-scale}\\
r^z_p(q_t)&=\frac{1}{c^z_p}\max_{0\leq i\leq K-p}\norm{\Delta^p\zh_i}.\label{eq:state-score}
\end{align}
\end{subequations}
The state-variation score is $\rstate(q_t)=\max_{p\in\{1,2,3\}}r^z_p(q_t)$. The factor 1.25 is a fixed heuristic margin. These are differences of standardized sample-index coordinates, not geometric curvature. Without restoring coordinate units and powers of the sampling interval, they are not physical velocity, acceleration, or jerk.

For commands, $V^a_p$ contains $\norm{\Delta^p\at_i^{(e)}}$ over the training episodes and $0\leq i\leq T_e-1-p$. For $p\in\{1,2\}$,
\begin{subequations}\label{eq:command}
\begin{align}
c^a_p&=1.25\,\Qlin_{0.995}(V^a_p)+\epsilon,\label{eq:command-scale}\\
r^a_p(q_t)&=\frac{1}{c^a_p}\max_{0\leq i\leq K-1-p}\norm{\Delta^p\at_i}.\label{eq:command-score}
\end{align}
\end{subequations}
The command-variation score is $\rcmd(q_t)=\max_{p\in\{1,2\}}r^a_p(q_t)$. In PushT, $a_i$ is a positional goal for the pusher, so command differences are not actuator velocity or effort. Appendix~\ref{app:proof} shows that the all-pairs displacement ratio cannot change a maximum that already contains $r^z_1$.

\subsection{Transition Residual and Ensemble Spread}
For each proposal transition $i=0,\ldots,K-1$, an ensemble predicts the next standardized state from $(\zh_i,\at_i)$. Member $f_m:\R^{d_z}\times\R^{d_a}\to\R^{d_z}$ predicts a state increment. With $M=5$ members,
\begin{align}
\zt^{(m)}_{i+1}&=\zh_i+f_m(\zh_i,\at_i),\label{eq:member}\\
\mu_{i,j}&=\frac{1}{M}\sum_{m=1}^{M}\zt^{(m)}_{i+1,j},\label{eq:mean}\\
\sigma_{i,j}&=\sqrt{\frac{1}{M-1}\sum_{m=1}^{M}(\zt^{(m)}_{i+1,j}-\mu_{i,j})^2}.
\label{eq:spread}
\end{align}
Thus $\mu_{i,j}$ is the ensemble mean and $\sigma_{i,j}$ the sample standard deviation for coordinate $j=1,\ldots,d_z$; $m=1,\ldots,M$ indexes ensemble members.

Set $\lambda=0.03$ in standardized-coordinate units and define the transition-level spread-scaled residual
\begin{equation}
e^{\sr}_i(q)=\sqrt{\frac{1}{d_z}\sum_{j=1}^{d_z}
\left(\frac{\zh_{i+1,j}-\mu_{i,j}}{\sigma_{i,j}+\lambda}\right)^2}.
\label{eq:sr}
\end{equation}
Let $q^{\sr}_1,\ldots,q^{\sr}_{B_{\sr}}$ be sampled nominal calibration windows and collect all their transition-level values in
$V_{\sr}=\{e^{\sr}_i(q^{\sr}_n):1\leq n\leq B_{\sr},\,0\leq i<K\}$.
In this study $B_{\sr}=600$ and $K=32$, giving 19,200 values. Set
\begin{equation}
c_{\sr}=1.20\,\Qlin_{0.995}(V_{\sr})+\epsilon.
\label{eq:csr}
\end{equation}
The three window-level statistics are
\begin{align}
\dRMS(q_t)&=\max_{0\leq i<K}\sqrt{\frac{1}{d_z}\sum_{j=1}^{d_z}(\zh_{i+1,j}-\mu_{i,j})^2},\label{eq:drms}\\
\rspread(q_t)&=\frac{1}{c_{\sr}}\max_{0\leq i<K}e^{\sr}_i(q_t),\label{eq:rspread}\\
u(q_t)&=\max_{0\leq i<K}\sqrt{\frac{1}{d_z}\sum_{j=1}^{d_z}\sigma_{i,j}^2}.
\label{eq:u}
\end{align}
The score $\dRMS$ measures disagreement with the proposed next state; $\rspread$ additionally scales each coordinate by ensemble spread plus $\lambda$. The quantity $u$ measures ensemble spread, not calibrated predictive uncertainty. The scale $c_{\sr}$ is neither a plant bound nor a window-level trigger threshold. Calibration windows and their transitions can be dependent, so 19,200 is a descriptive count.

\subsection{Trigger Statistic and Diagnostic Vector}
\label{sec:trigger}
\emph{One score decides whether to trigger, while the full vector records what the monitor observed}. Retaining several channels does not require taking maximum for the decision.

\textbf{Choose the scalar score.}
Let $d(q)\in\R$ denote the chosen score, with larger values indicating greater disagreement or variation. For example, a monitor can use $\dRMS$ alone. The heterogeneous comparator is
\begin{equation}
\dmax(q_t)=\max\{\rstate(q_t),\rcmd(q_t),\rspread(q_t)\}.
\label{eq:dmax}
\end{equation}

\textbf{Calibrate a window-level threshold.}
For $B$ nominal calibration windows $q^{\calib}_1,\ldots,q^{\calib}_B$, compute $s_b=d(q^{\calib}_b)$ and sort these scores as $s_{(1)}\leq\cdots\leq s_{(B)}$. Parentheses denote rank. Define
\begin{equation}
k=1+\lceil0.95(B-1)\rceil,\qquad \tau_d=s_{(k)},
\label{eq:tau}
\end{equation}
where $\lceil\cdot\rceil$ rounds upward to the next integer. With $B=600$, $k=571$: the threshold is simply the 571st smallest calibration score. The output is
\begin{equation}
\operatorname{trigger}(q_t)=
\begin{cases}
1,&d(q_t)>\tau_d,\\
0,&d(q_t)\leq\tau_d.
\end{cases}
\label{eq:trigger}
\end{equation}
This describes the calibration sample. Windows may share an episode, and selecting the best score from the reported test comparison is not an independent calibration step.

\textbf{Retain all channels.}
Regardless of which scalar triggers, store the eight-component vector
\begin{equation}
g(q_t)=\big[r^z_1,r^z_2,r^z_3,r^a_1,r^a_2,\dRMS,\rspread,u\big]^{\mathsf T},
\label{eq:log}
\end{equation}
with every component evaluated at $q_t$. Its first five entries describe state and command variation; the last three describe transition disagreement and ensemble spread. A high residual with modest variation therefore remains distinguishable from a large sequence difference. This is diagnostic information, not identification of a perturbation's physical cause. Likewise, a zero trigger does not certify safety or automatically authorize execution. The window threshold $\tau_d$ is separate from the normalization-derived $S=1$ threshold used in the additional fixed-run plots in Section~V-D.
\section{PushT Case Study}
\label{sec:study}
\textbf{Data and split.}
We use recorded demonstrations from the LeRobot PushT dataset~\cite{lerobot,diffusion}. The field \texttt{observation.state} is the planar position of the circular pusher and \texttt{action} is its planar position goal, so $d_z=d_a=2$. The environment applies each goal through a local PD controller at a 10\,Hz control rate~\cite{diffusion}. Each example contains $K=32$ transitions, 33 pusher positions and 32 goals, covering 3.2\,s. These recorded windows are offline surrogates for decoded proposals.

The scripts read at most 80,000 frames. For $N$ retained episodes, $\lfloor0.70N\rfloor$ are assigned to training, $\lfloor0.15N\rfloor$ to calibration, and the remainder to test; $\lfloor\cdot\rfloor$ denotes rounding downward. Training and evaluation first sample an episode uniformly and then a valid start index uniformly, both with replacement.

\begin{figure}[t]
\centering
\includegraphics[width=\columnwidth]{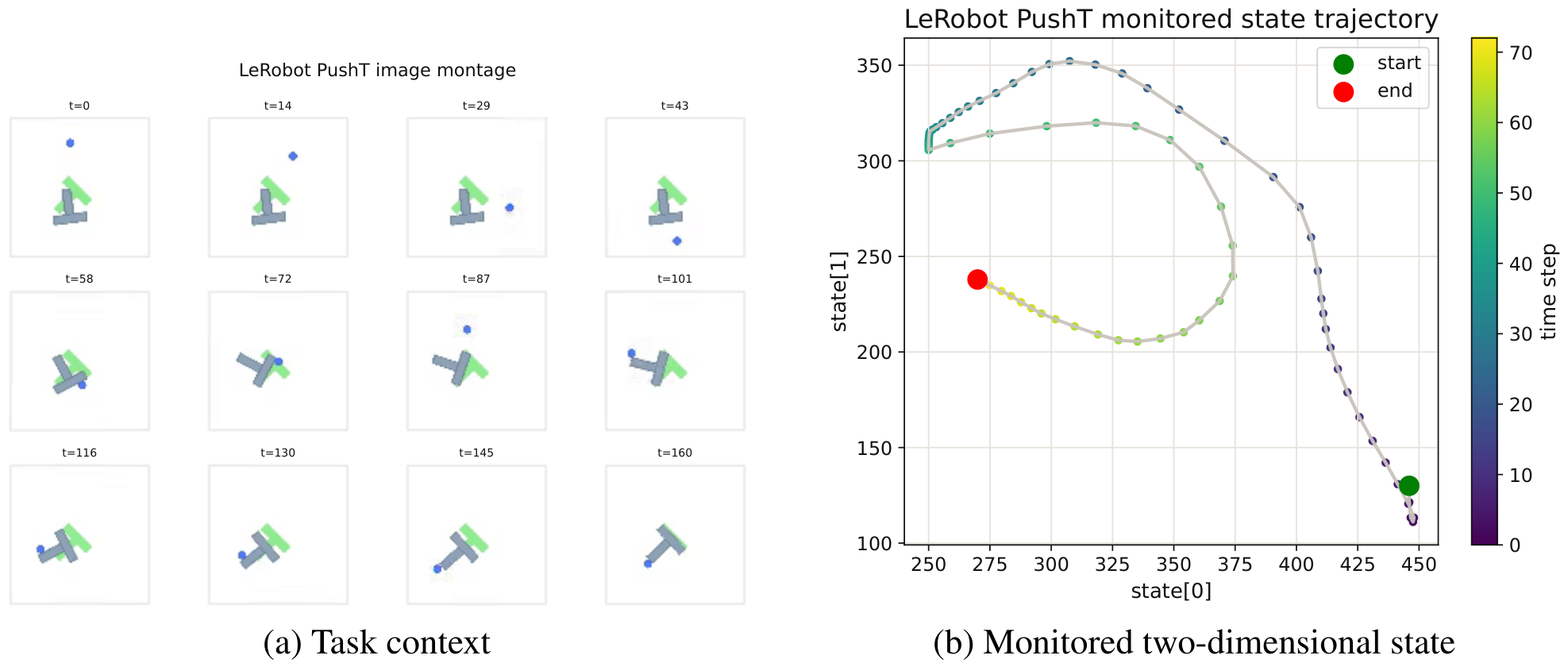}
\caption{One LeRobot PushT demonstration. All predictors and diagnostics use only the two-dimensional pusher position (right) and the recorded position goals.}
\label{fig:pusht}
\end{figure}

\textbf{Predictors.}
Three standard multilayer-perceptron baselines predict the pusher coordinates. The first is an ensemble of five independently initialized one-step delta predictors conditioned on the current position and current goal; each has four hidden layers of width 256 with SiLU activations. The history-conditioned predictor receives four recent positions, four preceding goals, and the current goal. A direct 32-step predictor has five hidden layers of width 384 and a horizon-weighted mean-squared-error loss whose weights increase linearly from 1.0 to 1.5. The three models receive 30,000, 20,000, and 18,000 AdamW updates, respectively, with batch size 256, learning rate $3\times10^{-4}$, weight decay $10^{-5}$, and gradient-norm clipping at 5.

\textbf{Synthetic perturbations.}
The fixed run draws 700 nominal test windows with replacement and independently resamples 175 test windows for each of 30 combinations of perturbation family and parameter value. There are six operators (Table~\ref{tab:operators}) and five values $\rho\in\{0.25,0.5,1,2,4\}$, producing $6\times5\times175=5{,}250$ transformed windows. The sampled base windows are not a paired bank shared across parameter values.

\begin{table}[t]
\caption{Synthetic perturbation operators.}
\label{tab:operators}
\centering
\small
\begin{tabularx}{\columnwidth}{@{}lL{2.05cm}Y@{}}
\toprule
ID & Family / stream & Synthetic edit\\
\midrule
P1 & Smooth pulse / position & Eight-sample half-sine displacement; six interior offsets are nonzero.\\[3pt]
P2 & Delayed suffix / position & Copy an earlier suffix with delay $\max\{1,\operatorname{round}(1+\rho)\}$.\\[3pt]
P3 & Compressed segment / position & Interpolate eight entries at rate $1+0.35\rho$, with endpoint clamping.\\[3pt]
P4 & Increment rotation / position & Rotate eight standardized displacements by $\min\{\pi,0.35\rho\}$; reconstruct the suffix.\\[3pt]
P5 & Reordered segment / goal & Reverse six standardized goals and multiply by $1+0.25\rho$.\\[3pt]
P6 & Shifted segment / goal & Add $2.5\rho c^a_1v$ to six standardized goals; $v$ is a sampled direction.\\
\bottomrule
\end{tabularx}
\end{table}

Every operator changes exactly one member of the recorded state--action pair while keeping the other fixed. The benchmark therefore directly tests cross-stream inconsistency and is structurally aligned with an action-conditioned transition residual. A positive label records only that an operator was applied; neither class is a physical-feasibility label. The 5,250 transformed windows are evaluated samples, not 5,250 independent episodes.

\textbf{Evaluation measures.}
Let $d^+_p$, $p=1,\ldots,P$, be transformed-window scores and $d^-_q$, $q=1,\ldots,Q$, nominal-window scores. Set $c_{pq}=1$, $1/2$, or $0$ according as $d^+_p>d^-_q$, $d^+_p=d^-_q$, or $d^+_p<d^-_q$. The empirical area under the receiver-operating-characteristic curve is
\begin{equation}
\widehat{\operatorname{AUC}}(d)=\frac{1}{PQ}\sum_{p=1}^{P}\sum_{q=1}^{Q}c_{pq}.
\label{eq:auc}
\end{equation}
Here $P=5{,}250$ and $Q=700$; ties receive half credit. Each family--parameter combination has equal weight in the pooled transformed class. We also report average precision (AP), with transformed windows as positives. The positive prevalence is $5{,}250/5{,}950\approx0.882$, which is the prevalence reference for interpreting AP. These ranking measures aggregate over score thresholds; they do not give the operating characteristics at the particular $\tau_d$ in~\eqref{eq:tau}.
\section{Results}
\label{sec:results}
The results address three different questions: which score ranks transformed windows most reliably, how accurately the predictors reproduce a recorded rollout, and how an existing fixed threshold responds across perturbation families.

\subsection{Ranking Nominal and Transformed Windows}
Table~\ref{tab:results} reports the fixed-run point estimates. The transition-RMSE baseline has the highest ROC AUC, 0.982, followed by the spread-scaled residual at 0.972. Their high ranking performance is consistent with the benchmark construction: changing one stream while freezing its paired stream directly introduces disagreement that an action-conditioned predictor can detect.

\begin{table}[t]
\caption{Fixed-run ranking performance.}
\label{tab:results}
\centering
\begin{tabularx}{\columnwidth}{@{}Yrr@{}}
\toprule
Score & ROC AUC & AP\\
\midrule
Uncertainty baseline & 0.828 & 0.968\\
Transition-RMSE baseline & \textbf{0.982} & \textbf{0.997}\\
Spread-scaled residual $\rspread$ & 0.972 & 0.995\\
State-difference score $\rstate$ & 0.592 & 0.901\\
Heterogeneous maximum $\dmax$ & 0.957 & 0.993\\
\bottomrule
\end{tabularx}
\end{table}

\begin{figure}[t]
\centering
\input{figures/rollout_means.tex}
\caption{Mean 32-step rollout RMSE in standardized pusher coordinates, from the fixed-run numerical summary. Mean values only; lower is better.}
\label{fig:rollout}
\end{figure}

\begin{figure*}[t]
\centering
\input{figures/family_diagnostics.tex}
\caption{Fixed-run diagnostics at the normalization-derived threshold. (a) Composite-score distributions for 700 nominal windows and 875 transformed windows per family; the dashed line is $S=1$. (b) Fraction of windows with $S>1$, with 175 sampled windows per cell. P1--P6 are the operators in Table~\ref{tab:operators}. The displayed threshold is normalization-derived and is not the calibrated window threshold $\tau_d$.}
\label{fig:families}
\end{figure*}

The state-difference score has AUC 0.592, substantially below the transition-based scores. It inspects the shape of the position sequence rather than agreement between that sequence and the commands. In particular, the goal-edit operators P5 and P6 leave positions unchanged, so $\rstate$ is exactly unchanged for each edited window relative to its own base window. The independently sampled nominal class need not contain that same base window.

AP requires separate interpretation. For example, $\rstate$ has AP 0.901 despite its much weaker AUC.

\subsection{What Heterogeneous Aggregation Changes}
The heterogeneous maximum reaches AUC 0.957 and AP 0.993. Its AUC is 0.025 below transition RMSE, 0.015 below the spread-scaled residual, and 0.365 above state differences. Thus adding channels improves over the weak state-only comparator in this run, but does not outperform the strongest transition-based comparator.

Although $\dmax(q)\geq\rspread(q)$ for every proposal, this pointwise inequality does not imply a higher AUC. An auxiliary variation channel can raise a nominal window's score more than a transformed window's score and thereby change their ordering. Moreover, dividing channels by empirical reference scales does not calibrate their relative ability to separate the two classes. These observations explain why keeping a channel for inspection need not justify including it in the trigger. They do not identify which channel caused each observed ranking change; that attribution requires the per-window score vectors.

The comparison therefore supports a restrained engineering choice: compare candidate trigger scores on the intended detection task, and retain the remaining channels in $g(q)$ rather than assuming that a maximum is beneficial.

\subsection{Predictor Rollout Accuracy Is a Separate Outcome}
All predictors use the standardized coordinates in~\eqref{eq:standardize}. For a $K$-step prediction $\zt^{\mathrm{pred}}_{1:K}$ and recorded target $\zt^{\mathrm{obs}}_{1:K}$, the window-level rollout error is
\begin{equation}
\operatorname{RMSE}_K=\sqrt{\frac{1}{Kd_z}\sum_{k=1}^{K}\norm{\zt^{\mathrm{pred}}_k-\zt^{\mathrm{obs}}_k}^{2}}.
\label{eq:rollout}
\end{equation}
The superscripts $\mathrm{pred}$ and $\mathrm{obs}$ distinguish predicted and recorded positions.

The three predictor architectures are described in Section~\ref{sec:study} (Predictors); their maps and rollout recursion are given in Supplementary Material, Section~S1, equations~(S1)--(S3). Fig.~\ref{fig:rollout} shows the reported mean 32-step rollout RMSE: 0.0100 for the current-state-and-action ensemble rollout, 0.0022 for the history-conditioned predictor, and 0.0226 for the direct predictor. The coordinates are standardized using the training statistics, so these values are not distances in the workspace.

Rollout RMSE in~\eqref{eq:rollout} averages predictive error over the horizon and coordinates. In contrast, $\dRMS$ in~\eqref{eq:drms} takes the maximum one-step disagreement with a proposed sequence. The former evaluates a predictor; the latter ranks proposals against a predictor.

\subsection{Family and Parameter Diagnostics at a Fixed Threshold}
Fig.~\ref{fig:families} shows fixed-run score distributions and family--parameter trigger fractions. It uses the implementation's composite score $S$ and the normalization-derived threshold $S>1$, not the nominal-window threshold $\tau_d$ of Section~III-C.

The plots reveal heterogeneity hidden by a pooled score: the small-parameter pulse, compression, and rotation edits have visibly lower trigger fractions than the goal edits. The compression response is not uniformly increasing across the displayed parameter values. The parameter $\rho$ changes a different algebraic operation in each row, so equal $\rho$ does not mean equal physical difficulty.

Each heatmap cell summarizes 175 resampled windows. Since base windows were independently resampled across parameter values, differences between cells combine the edit's effect with sampling variation.
\section{Limitations}
\label{sec:limitations}
Only planar pusher position is observed as state. Synthetic one-stream edits favor transition residuals; AP reflects the artificial 88.2\% positive prevalence. Results use one split and initialization set, with dependent windows and no per-window records for interval estimation or independent re-evaluation. Calibrated-threshold test false-trigger rates, closed-loop benefits, and diagnostic-cause accuracy are not established. Composition consistency remains untested.

\section{Conclusion}
We formalized a restricted prediction--control interface, separated empirical variation from learned-transition diagnostics, and established the redundancy of an all-pairs displacement ratio within a corresponding maximum. Six explicit PushT edits provided a controlled test of proposal inconsistency. In the fixed run, transition RMSE (AUC 0.982) and the spread-scaled residual (0.972) both outperformed the heterogeneous maximum (0.957), while state differences alone were substantially weaker (0.592).
\appendices
\section{Redundancy of the Pairwise Displacement Ratio}
\label{app:proof}
Let $c_1>0$ be the one-step reference scale and $\eta\geq0$ a denominator offset. For $\zt_0,\ldots,\zt_K$, define
\begin{equation}
\begin{split}
r_1&=\max_{0\leq i<K}\frac{\norm{\zt_{i+1}-\zt_i}}{c_1},\\
r_{\mathrm{pair}}^{(\eta)}&=\max_{0\leq i<j\leq K}
\frac{\norm{\zt_j-\zt_i}}{(j-i)c_1+\eta}.
\end{split}
\label{eq:pair}
\end{equation}
In the implementation $c_1=c^z_1$ and $\eta=0.1c_1$. The triangle inequality gives, for every $i<j$,
\begin{equation}
\norm{\zt_j-\zt_i}\leq\sum_{s=i}^{j-1}\norm{\zt_{s+1}-\zt_s}
\leq(j-i)c_1r_1.
\label{eq:triangle}
\end{equation}
Consequently,
\begin{equation}
r_{\mathrm{pair}}^{(\eta)}\leq\max_{i<j}\frac{(j-i)c_1r_1}{(j-i)c_1+\eta}\leq r_1.
\label{eq:redundant}
\end{equation}
Thus $\max\{r_1,r_{\mathrm{pair}}^{(\eta)}\}=r_1$: the pairwise term cannot change a max composite that already contains the one-step score, nor can it be the sole channel to cross a common threshold. It may induce a different ranking when used alone.

\end{document}

%% file: figures/interface.tex
\begin{tikzpicture}[x=1cm,y=1cm,font=\small,
  box/.style={draw,rounded corners=2pt,align=center,inner sep=5pt},
  flow/.style={-{Stealth[length=2mm,width=1.5mm]},line width=.75pt}]
  \node[box,text width=2.05cm,minimum height=1.8cm] (model) at (1.13,0)
    {\textbf{Predictive model}\\[3pt]Policy / VLA\\World model};
  \node[box,text width=2.25cm,minimum height=1.8cm] (proposal) at (4.22,0)
    {\textbf{Decoded proposal}\\[3pt]States $\hat z_{0:K}$\\Commands $a_{0:K-1}$};
  \node[box,text width=5.0cm,minimum height=4.25cm,fill=black!3] (monitor) at (8.88,0) {};
  \node[font=\small\bfseries] at (8.88,1.78) {Proposal diagnostic monitor};
  \node[box,text width=4.5cm,minimum height=.8cm,inner sep=3pt,fill=white] at (8.88,.95)
    {Empirical state and command variation};
  \node[box,text width=4.5cm,minimum height=.8cm,inner sep=3pt,fill=white] at (8.88,-.15)
    {Learned-transition residual\\and ensemble spread};
  \node[box,dashed,text width=4.5cm,minimum height=.8cm,inner sep=3pt,fill=white] at (8.88,-1.27)
    {Direct / composed forecasts: optional};
  \node[box,text width=3.1cm,minimum height=1.1cm] (trigger) at (14.3,.79)
    {\textbf{Trigger}\\$d(q_t)>\tau_d$};
  \node[box,text width=3.1cm,minimum height=1.1cm] (log) at (14.3,-.79)
    {\textbf{Diagnostic log}\\All channels $g(q_t)$};
  \draw[flow] (model.east) -- (proposal.west);
  \draw[flow] (proposal.east) -- (monitor.west);
  \draw[flow] (monitor.east |- trigger.west) -- (trigger.west);
  \draw[flow] (monitor.east |- log.west) -- (log.west);
\end{tikzpicture}

%% file: figures/rollout_means.tex
\begin{tikzpicture}
\begin{axis}[
 width=.91\columnwidth,height=4.6cm,
 ybar,bar width=20pt,
 ymin=0,ymax=.027,
 symbolic x coords={Ensemble,History,Direct},xtick=data,
 ylabel={Mean rollout RMSE},
 ytick={0,.005,.010,.015,.020,.025},
 scaled y ticks=false,
 yticklabel style={/pgf/number format/fixed,/pgf/number format/precision=3},
 tick label style={font=\footnotesize},label style={font=\small},
 enlarge x limits=.3,
 nodes near coords,
 every node near coord/.append style={font=\footnotesize,/pgf/number format/fixed,/pgf/number format/precision=4},
 axis lines*=left,ymajorgrids=true,grid style={densely dotted},
]
\addplot[draw=black,fill=black!18] coordinates {(Ensemble,.0100) (History,.0022) (Direct,.0226)};
\end{axis}
\end{tikzpicture}

%% file: figures/family_diagnostics.tex
\begin{tikzpicture}[x=1cm,y=1cm,font=\small]
\node[anchor=north west,font=\small\bfseries] at (0,0) {(a) Composite-score distributions};
\node[anchor=north west,inner sep=0] at (.2,-.6)
 {\includegraphics[width=7.7cm,height=2.7cm]{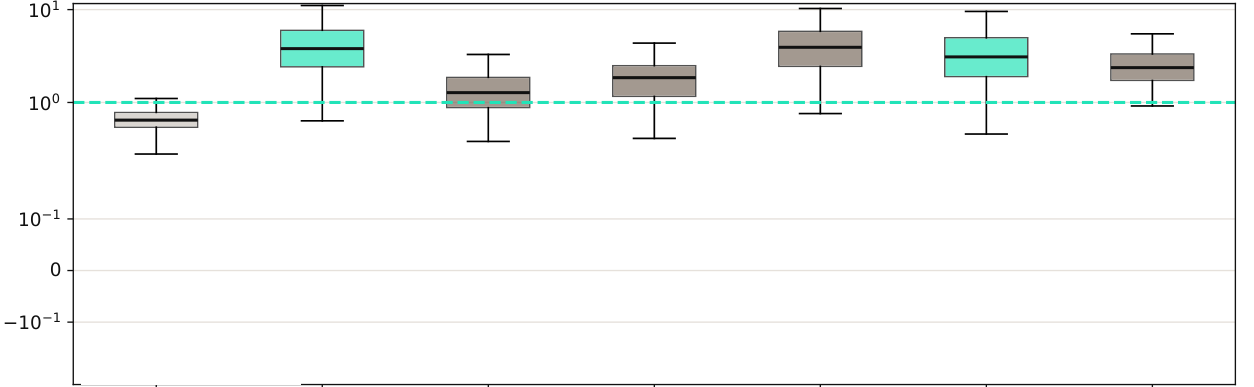}};
\node[rotate=90,font=\footnotesize] at (-.12,-1.93) {Score $S$};
\foreach \x/\lab in {1.06/Nominal,2.13/P1,3.20/P2,4.27/P3,5.34/P4,6.41/P5,7.48/P6}
  \node[font=\footnotesize,anchor=north] at (\x,-3.38) {\lab};
\node[font=\footnotesize,anchor=north] at (4,-3.87) {Dashed line: $S=1$ (not $\tau_d$)};
\node[anchor=north west,font=\small\bfseries] at (9,0) {(b) Trigger fractions at $S>1$};
\node[anchor=north west,inner sep=0] at (9.35,-.5)
 {\includegraphics[width=5.10cm,height=3.9cm]{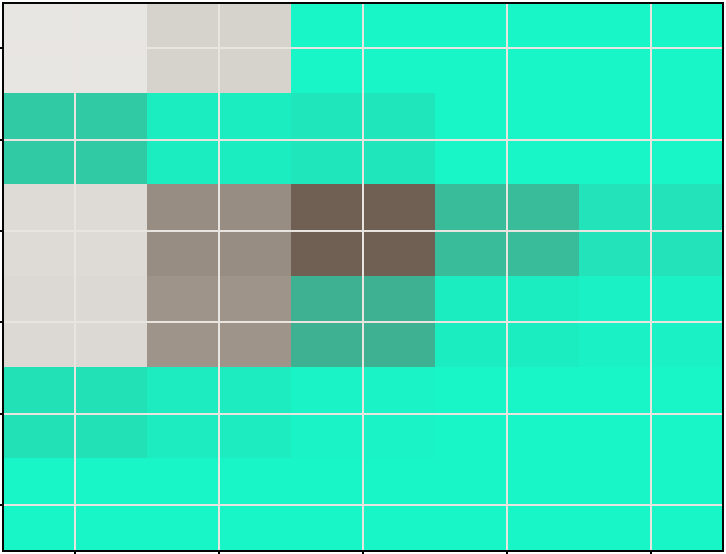}};
\foreach \y/\lab in {-.825/P1,-1.475/P2,-2.125/P3,-2.775/P4,-3.425/P5,-4.075/P6}
 \node[font=\footnotesize,anchor=east] at (9.23,\y) {\lab};
\foreach \x/\lab in {9.86/0.25,10.88/0.5,11.90/1,12.92/2,13.94/4}
 \node[font=\footnotesize,anchor=north] at (\x,-4.46) {$\lab$};
\node[font=\footnotesize,anchor=north] at (11.9,-4.91) {Operator parameter $\rho$};
\node[anchor=north west,inner sep=0] at (14.75,-.405)
 {\includegraphics[height=4.06cm]{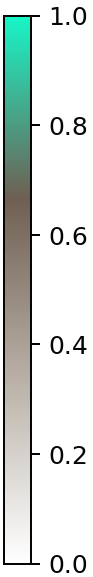}};
\node[rotate=90,font=\footnotesize] at (15.77,-2.45) {Trigger fraction};
\end{tikzpicture}

%% file: main.bbl
\begin{thebibliography}{10}
\providecommand{\url}[1]{#1}
\csname url@samestyle\endcsname
\providecommand{\newblock}{\relax}
\providecommand{\bibinfo}[2]{#2}
\providecommand{\BIBentrySTDinterwordspacing}{\spaceskip=0pt\relax}
\providecommand{\BIBentryALTinterwordstretchfactor}{4}
\providecommand{\BIBentryALTinterwordspacing}{\spaceskip=\fontdimen2\font plus
\BIBentryALTinterwordstretchfactor\fontdimen3\font minus
  \fontdimen4\font\relax}
\providecommand{\BIBforeignlanguage}[2]{{%
\expandafter\ifx\csname l@#1\endcsname\relax
\typeout{** WARNING: IEEEtran.bst: No hyphenation pattern has been}%
\typeout{** loaded for the language `#1'. Using the pattern for}%
\typeout{** the default language instead.}%
\else
\language=\csname l@#1\endcsname
\fi
#2}}
\providecommand{\BIBdecl}{\relax}
\BIBdecl

\bibitem{rt1}
A.~Brohan, N.~Brown, J.~Carbajal, Y.~Chebotar, J.~Dabis, C.~Finn,
  K.~Gopalakrishnan, K.~Hausman, A.~Herzog, J.~Hsu \emph{et~al.}, ``{RT-1}:
  Robotics transformer for real-world control at scale,'' arXiv:2212.06817,
  2022.

\bibitem{rt2}
A.~Brohan, N.~Brown, J.~Carbajal, Y.~Chebotar, X.~Chen, K.~Choromanski,
  T.~Ding, D.~Driess, A.~Dubey, C.~Finn \emph{et~al.}, ``{RT-2}:
  Vision-language-action models transfer web knowledge to robotic control,''
  arXiv:2307.15818, 2023.

\bibitem{openx}
{Open X-Embodiment Collaboration}, ``Open {X}-embodiment: Robotic learning
  datasets and {RT-X} models,'' arXiv:2310.08864, 2023.

\bibitem{octo}
{Octo Model Team}, D.~Ghosh, H.~Walke, K.~Pertsch, K.~Black, O.~Mees,
  S.~Dasari, J.~Hejna, T.~Kreiman, C.~Xu \emph{et~al.}, ``Octo: An open-source
  generalist robot policy,'' arXiv:2405.12213, 2024.

\bibitem{openvla}
M.~J. Kim, K.~Pertsch, S.~Karamcheti, T.~Xiao, A.~Balakrishna, S.~Nair,
  R.~Rafailov, E.~Foster, G.~Lam, P.~Sanketi \emph{et~al.}, ``{OpenVLA}: An
  open-source vision-language-action model,'' arXiv:2406.09246, 2024.

\bibitem{pi0}
K.~Black, N.~Brown, D.~Driess, A.~Esmail, M.~Equi, C.~Finn, N.~Fusai, L.~Groom,
  K.~Hausman, B.~Ichter \emph{et~al.}, ``$\pi_0$: A vision-language-action flow
  model for general robot control,'' arXiv:2410.24164, 2024.

\bibitem{or2022hybrid}
B.~Or and I.~Klein, ``A hybrid model and learning-based adaptive navigation
  filter,'' \emph{IEEE Transactions on Instrumentation and Measurement},
  vol.~71, 2022, art. no.~2516311,
  {doi:}~\href{https://doi.org/10.1109/TIM.2022.3197775}{\nolinkurl{10.1109/TIM.2022.3197775}}.

\bibitem{or2023trajectory}
------, ``Learning vehicle trajectory uncertainty,'' \emph{Engineering
  Applications of Artificial Intelligence}, vol. 122, 2023, art. no.~106101,
  {doi:}~\href{https://doi.org/10.1016/j.engappai.2023.106101}{\nolinkurl{10.1016/j.engappai.2023.106101}}.

\bibitem{freydin2022speed}
M.~Freydin and B.~Or, ``Learning car speed using inertial sensors for dead
  reckoning navigation,'' \emph{IEEE Sensors Letters}, vol.~6, no.~9, 2022,
  art. no.~6003104,
  {doi:}~\href{https://doi.org/10.1109/LSENS.2022.3201731}{\nolinkurl{10.1109/LSENS.2022.3201731}}.

\bibitem{ames}
A.~D. Ames, S.~Coogan, M.~Egerstedt, G.~Notomista, K.~Sreenath, and P.~Tabuada,
  ``Control barrier functions: Theory and applications,'' in \emph{Proc.
  European Control Conf. (ECC)}, 2019, pp. 3420--3431.

\bibitem{wabersich}
K.~P. Wabersich and M.~N. Zeilinger, ``A predictive safety filter for
  learning-based control of constrained nonlinear dynamical systems,''
  \emph{Automatica}, vol. 129, p. 109597, 2021.

\bibitem{hsu}
K.-C. Hsu, H.~Hu, and J.~F. Fisac, ``The safety filter: A unified view of
  safety-critical control in autonomous systems,'' \emph{Annual Review of
  Control, Robotics, and Autonomous Systems}, vol.~7, pp. 47--72, 2024.

\bibitem{simplex}
D.~Seto, B.~H. Krogh, L.~Sha, and A.~Chutinan, ``The simplex architecture for
  safe online control system upgrades,'' in \emph{Proc. American Control
  Conf.}, 1998, pp. 3504--3508.

\bibitem{shielding}
\BIBentryALTinterwordspacing
M.~Alshiekh, R.~Bloem, R.~Ehlers, B.~K{"o}nighofer, S.~Niekum, and U.~Topcu,
  ``Safe reinforcement learning via shielding,'' in \emph{Proc. AAAI Conf.
  Artificial Intelligence}, vol.~32, 2018. [Online]. Available:
  \url{https://arxiv.org/abs/1708.08611}
\BIBentrySTDinterwordspacing

\bibitem{hobbs}
K.~L. Hobbs, M.~L. Mote, M.~Abate, S.~Coogan, and E.~Feron, ``Run time
  assurance for safety-critical systems: An introduction to safety filtering
  approaches for complex control systems,'' \emph{IEEE Control Systems
  Magazine}, vol.~43, no.~2, pp. 28--65, 2023.

\bibitem{commonroad}
C.~Pek, V.~Rusinov, S.~Manzinger, M.~C. {\"U}ste, and M.~Althoff,
  ``{CommonRoad} drivability checker: Simplifying the development and
  validation of motion planning algorithms,'' in \emph{Proc. IEEE Intelligent
  Vehicles Symposium (IV)}, 2020, pp. 1013--1020.

\bibitem{forceopt}
K.~Chakraborty, Z.~Feng, S.~Veer, A.~Sharma, W.~Ding, S.~Topan, B.~Ivanovic,
  M.~Pavone, and S.~Bansal, ``Safety evaluation of motion plans using
  trajectory predictors as forward reachable set estimators,'' \emph{IEEE
  Robotics and Automation Letters}, vol.~11, no.~3, pp. 3262--3269, 2026.

\bibitem{shikhman}
\BIBentryALTinterwordspacing
L.~J. Shikhman, ``Semigroup consistency as a diagnostic for learned physics
  simulators,'' in \emph{AI4Physics Workshop, International Conf. Machine
  Learning}, 2026. [Online]. Available: \url{https://arxiv.org/abs/2605.26324}
\BIBentrySTDinterwordspacing

\bibitem{diffusion}
C.~Chi, S.~Feng, Y.~Du, Z.~Xu, E.~Cousineau, B.~C.~M. Burchfiel, and S.~Song,
  ``Diffusion policy: Visuomotor policy learning via action diffusion,'' in
  \emph{Proc. Robotics: Science and Systems}, 2023.

\bibitem{act}
T.~Z. Zhao, V.~Kumar, S.~Levine, and C.~Finn, ``Learning fine-grained bimanual
  manipulation with low-cost hardware,'' arXiv:2304.13705, 2023.

\bibitem{pets}
K.~Chua, R.~Calandra, R.~McAllister, and S.~Levine, ``Deep reinforcement
  learning in a handful of trials using probabilistic dynamics models,'' in
  \emph{Advances in Neural Information Processing Systems}, 2018.

\bibitem{mbpo}
M.~Janner, J.~Fu, M.~Zhang, and S.~Levine, ``When to trust your model:
  Model-based policy optimization,'' in \emph{Advances in Neural Information
  Processing Systems}, 2019.

\bibitem{dreamer}
D.~Hafner, T.~Lillicrap, J.~Ba, and M.~Norouzi, ``Dream to control: Learning
  behaviors by latent imagination,'' in \emph{Proc. International Conf.
  Learning Representations}, 2020.

\bibitem{dreamerv3}
D.~Hafner, J.~Pasukonis, J.~Ba, and T.~Lillicrap, ``Mastering diverse domains
  through world models,'' arXiv:2301.04104, 2023.

\bibitem{revach2022kalmannet}
G.~Revach, N.~Shlezinger, X.~Ni, A.~L. Escoriza, R.~J.~G. van Sloun, and Y.~C.
  Eldar, ``{KalmanNet}: Neural network aided {Kalman} filtering for partially
  known dynamics,'' \emph{IEEE Transactions on Signal Processing}, vol.~70, pp.
  1532--1547, 2022,
  {doi:}~\href{https://doi.org/10.1109/TSP.2022.3158588}{\nolinkurl{10.1109/TSP.2022.3158588}}.

\bibitem{liu2020tlio}
W.~Liu, D.~Caruso, E.~Ilg, J.~Dong, A.~I. Mourikis, K.~Daniilidis, V.~Kumar,
  and J.~Engel, ``{TLIO}: Tight learned inertial odometry,'' \emph{IEEE
  Robotics and Automation Letters}, vol.~5, no.~4, pp. 5653--5660, 2020,
  {doi:}~\href{https://doi.org/10.1109/LRA.2020.3007421}{\nolinkurl{10.1109/LRA.2020.3007421}}.

\bibitem{safeRL}
J.~Garc{\'i}a and F.~Fern{\'a}ndez, ``A comprehensive survey on safe
  reinforcement learning,'' \emph{Journal of Machine Learning Research},
  vol.~16, no.~1, pp. 1437--1480, 2015.

\bibitem{ensembles}
B.~Lakshminarayanan, A.~Pritzel, and C.~Blundell, ``Simple and scalable
  predictive uncertainty estimation using deep ensembles,'' in \emph{Advances
  in Neural Information Processing Systems}, 2017.

\bibitem{lerobot}
R.~Cadene, S.~Aliberts, F.~Capuano, M.~Aractingi, A.~Zouitine, P.~Kooijmans,
  J.~Choghari, M.~Russi, C.~Pascal, S.~Palma, M.~Shukor, J.~Moss, A.~Soare,
  D.~Aubakirova, Q.~Lhoest, Q.~Gallou{\'e}dec, and T.~Wolf, ``{LeRobot}: An
  open-source library for end-to-end robot learning,'' arXiv:2602.22818, 2026.

\end{thebibliography}
